\documentclass[11pt]{article}
\usepackage{coling2020}
\usepackage{times}
\usepackage{url}
\usepackage{latexsym}

\usepackage{amsmath}
\usepackage{amsmath}
\usepackage{tabularx}
\usepackage{amsfonts}
\usepackage{url}
\usepackage{epsfig}
\usepackage{multirow}
\usepackage{graphicx}
\usepackage{subcaption}
\usepackage{booktabs}
\usepackage{graphicx}
\usepackage{wrapfig}
\usepackage{todonotes}
\newcommand{\tabhead}[1]{\textbf{#1}}

\makeatletter
\newcommand\Label[1]{&\refstepcounter{equation}(\theequation)\ltx@label{#1}&}

\makeatother

\colingfinalcopy

\title{Seeing Both the Forest and the Trees: Multi-head Attention for Joint Classification on Different Compositional Levels}

\author{Miruna P\^{i}slar \\ 
        Dept. of Computer Science, \\ University of Cambridge, UK \\
        \\
        {\tt miruna.pislar@gmail.com}
        \And
        Marek Rei \\ 
        Dept. of Computing, Imperial College London \\ 
        The ALTA Institute, Computer Laboratory, \\ University of Cambridge, UK \\
        {\tt marek.rei@imperial.ac.uk}}
        
\author{
Miruna P\^{i}slar$^{\spadesuit}$ ~ Marek Rei$^\clubsuit{}^\diamondsuit$
 \\
$^\spadesuit$Dept. of Computer Science, University of Cambridge, UK\\
$^\clubsuit$Dept. of Computing, Imperial College London, UK \\
$^\diamondsuit$The ALTA Institute, Dept. of Computer Science, University of Cambridge, UK\\ 
{ \small \tt miruna.pislar@gmail.com, marek.rei@imperial.ac.uk}
}

\date{}

\begin{document}
\maketitle
\begin{abstract}
In natural languages, words are used in association to construct sentences. It is not words in isolation, but the appropriate combination of hierarchical structures that conveys the meaning of the whole sentence. Neural networks can capture expressive language features; however, insights into the link between words and sentences are difficult to acquire automatically. In this work, we design a deep neural network architecture that explicitly wires lower and higher linguistic components; we then evaluate its ability to perform the same task at different hierarchical levels. Settling on broad text classification tasks, we show that our model, MHAL, learns to simultaneously solve them at different levels of granularity by fluidly transferring knowledge between hierarchies. Using a multi-head attention mechanism to tie the representations between single words and full sentences, MHAL systematically outperforms equivalent models that are not incentivized towards developing compositional representations. Moreover, we demonstrate that, with the proposed architecture, the sentence information flows naturally to individual words, allowing the model to behave like a sequence labeller (which is a lower, word-level task) even without any word supervision, in a zero-shot fashion.
\end{abstract}

\section{Introduction}
\label{sec:introduction}

\blfootnote{
    \hspace{-0.65cm}
    This work is licensed under a Creative Commons 
    Attribution 4.0 International License.
    License details:
    \url{http://creativecommons.org/licenses/by/4.0/}.
}

\blfootnote{
    \hspace{-0.65cm}
    Implementation and resources: \url{https://github.com/MirunaPislar/multi-head-attention-labeller}.
}

Compositional reasoning is fundamental in human cognition: we use it to interact with objects, take actions, reason about numbers, and move in space ~\cite{spelke}. This is also reflected in some aspects of the human language ~\cite{Wagner2011,Piantadosi,Sandler} since we use words and phrases in association to construct sentences. Consequently, there are lower linguistic components that act as building blocks for higher levels.

In this work, we focus on two levels of the compositional hierarchy -- words and sentences -- and ask the following question: are deep neural networks (DNNs) trained for a higher-level task (i.e., at the sentence level) able to pick up the features of the compounds needed to solve the same task but at a lower level, such as at the word level? Moreover, how are the different hierarchical levels interacting under varying supervision signals? To the best of our knowledge, very few studies have investigated the transferability of a task solution between compositional levels using DNNs in controlled experiments.

It has been shown that neural networks are universal function approximators ~\cite{hornik,Moshe_ufa}; they can perform arbitrary function combinations to learn expressive features. DNNs trained for language tasks are not an exception to this rule, and recent studies have shown their power in extracting linguistically-rich representations ~\cite{Mikolov,Bert}. However, when trained end-to-end, learning the connection between the different compositional levels is not trivial for these models. This is in part due to the vast syntactic and semantic complexity of natural language. There are also data limitations on most tasks, resulting in networks picking up the noise and biases of the datasets. Crucially, DNNs trained to solve a task at a higher hierarchical level are usually treated as black boxes with respect to the lower levels.

We propose a novel DNN design that stimulates the development of hierarchical connections. 
The architecture is based on a multi-head attention mechanism that ties the representations between single words and full sentences in a way that enables them to reinforce each other. 
The proposed multi-level architecture can be viewed as a sentence classifier, where each customized attention head is guided to behave like a sequence labeler, detecting one particular label on each token. Thus, it can simultaneously solve language tasks that are situated at different levels of granularity. Based on experiments, this architecture systematically outperforms equivalent models that focus only on one level. The token-level supervision explicitly teaches the classifier which areas it needs to focus on in each sentence, while the sentence-level objective provides a regularizing effect and encourages the model to return coherent sequence labeling predictions. Moreover, we show that the sentence-level information flows naturally to individual words, allowing the model to behave like a sequence labeler even when it does not receive any word-level supervision. Our model exhibits strong transfer capabilities, which we validated on three different tasks: sentiment analysis, named entity recognition, and grammatical error detection.

\section{Multi-head attention labeling (MHAL)}
\label{sec:mhal}

We describe an architecture that directly ties together the sentence and word representations for multi-class classification, incentivizing the model to make better use of the information on each level of granularity. In addition, we present several auxiliary objectives that guide this architecture towards useful hierarchical representations and better performance.

\subsection{Architecture}

Our model is based on a bidirectional long short-term memory (BiLSTM) that builds contextual vector representations for each word. These vectors are then passed through a multi-head attention mechanism, which predicts label distributions for both individual words and the whole sentence. Each attention head is incentivized to be predictive of a particular label, allowing the system to also assign labels to individual words while composing a sentence-level representation for sentence classification. 

The network takes as input a tokenized sentence of length $N$ and maps it to a sequence of vectors $[x_1, x_2, ..., x_N]$. Each vector $x_i$, corresponding to the $i^{th}$ token in a sentence, is the concatenation of its pre-trained GloVe word embedding $w_i$ ~\cite{glove} with its character-level representation $c_i$, similar to \newcite{lample-etal-2016-neural}. Passing each vector $x_i$ to a BiLSTM ~\cite{GravesAndSchmidhuber}, we obtain compact token representations $z_i$ by concatenating the hidden states from each direction at every time step and projecting these onto a joint feature space using a \textit{tanh} activation (Equations \ref{eq:lstm_word_1}-\ref{eq:lstm_word_2}).

This is followed by a multi-head attention mechanism with $H$ heads ~\cite{attentionAll}. By setting $H$ equal to the size of our token-level tagset, we can create a direct one-to-one correspondence between attention heads and possible token labels -- attention head $h \in \{1, 2, ..., H\}$ gets assigned to the $h$-th token-level label. 
For each attention head  we calculate keys, queries and values at every word position through a non-linear projection of $z_i$ (Equations \ref{eq:keys_1}-\ref{eq:values_2}). All the queries for a given attention head are then combined into a single vector through averaging, which will represent a query for the corresponding token-level label in the context of the given sentence (Equation \ref{eq:queries}).

\noindent\begin{minipage}{0.5\linewidth}
\begin{align}
\overrightarrow{z_i} & = \text{LSTM}(x_i, \overrightarrow{z_{i-1}}) \label{eq:lstm_word_1} \\
\overleftarrow{z_i} & = \text{LSTM}(x_i, \overleftarrow{z_{i+1}}) \\
z_i & = \tanh([\overrightarrow{z_i}; \overleftarrow{z_i}] W_z + b_z) \label{eq:lstm_word_2}
\end{align}

\end{minipage}%
\begin{minipage}{0.5\linewidth}
\begin{align}
k_{ih} & = \tanh(z_i W_{kh} + b_{kh}) \label{eq:keys_1} \\
q_{ih} & = \tanh(z_i W_{qh} + b_{qh}) \\
v_{ih} & = \tanh(z_i W_{vh} + b_{vh}) \label{eq:values_2}
\end{align}
\end{minipage}\par\vspace{\belowdisplayskip}

\noindent where $\overrightarrow{z_i}$ and $\overleftarrow{z_i}$ are LSTM hidden states in either direction; $W_z$, $W_{kh}$, $W_{qh}$ and $W_{vh}$ are weight matrices; and $b_z$, $b_{kh}$, $b_{qh}$ and $b_{vh}$ are bias vectors.

The unnormalized attention scores $a_{ih} \in \mathbb{R}^{1}$ are then calculated through a dot product between the query and the associated key for a particular token in position $i$ (Equation \ref{eq:attn_1}). Given the established correspondence between attention heads and token labels, this score now represents the model confidence that the token in position $i$ has label $h$. Therefore, we can predict the probability distribution over the token-level labels by normalizing $a_{ih}$ with a softmax function (Equation \ref{eq:token_out}). By concatenating the scores and normalizing them across the heads $h$, we get $\tilde{t}_{i} \in \mathbb{R}^{H}$, which we use as the token-level output from the model, for both optimization and evaluation as a token-level tagger.

\noindent\begin{minipage}[h]{0.5\textwidth}
\begin{align}
q_{h} & = \frac{1}{N} \sum_{i=1}^{N}{q_{ih}} \label{eq:queries} \\
a_{ih} & = q_{h} \cdot k_{ih} \label{eq:attn_1} \\
\tilde{t}_{ih} & = \frac{exp(a_{ih})}{\sum_{h'=1}^{N}{exp(a_{ih'})}} \label{eq:token_out}
\end{align}
\end{minipage}%
\begin{minipage}[h]{0.5\textwidth}
\begin{align}
\alpha_{ih} & = \frac{\sigma(a_{ih})}{{\sum_{j=1}^{N}{\sigma(a_{jh})}} \label{eq:norm_1}} \\
s_{h} & = \sum_{i=1}^{N}{\alpha_{ih} v_{ih}} \label{eq:sentence_rep} \\
o_h & = W_o \tanh(W_s s_h + b_s) + b_o \label{eq:attn_2}
\end{align}
\end{minipage}
\vspace{0.2cm}

\begin{wrapfigure}{R}{0.4\textwidth}
\centering
\includegraphics[width=0.38\columnwidth]{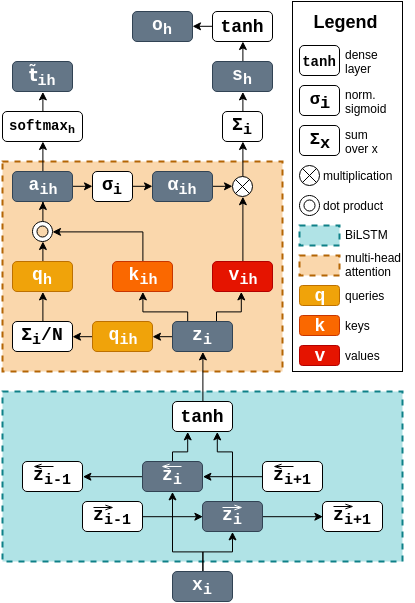}
\caption{Illustration of the MHAL architecture for one head $h$ only. We present the computations performed for the $i^{th}$ word in a sentence, mapped to its vector representation $x_i$.}\label{fig:architecture}
\end{wrapfigure}

Next, we use the same attention scores $a_{ih}$ to construct sentence-level representations of the input.
We apply a sigmoid activation ($\sigma$) and normalization to produce the normalized attention weights $\alpha_{ih} \in [0, 1]$ (Equation \ref{eq:norm_1}). 
Standard attention functions use a softmax activation, which is best suited for assigning most of the attention weight to a single token and effectively ignoring the rest. However, it is often necessary that higher-level representations pay attention to many different locations, or, in our case, to multiple tokens in a given sentence. By using the sigmoid instead of softmax, similar to \newcite{shen_and_lee_16}, the model will need to make separate decisions for each token, which in turn encourages the attention scores to behave more similarly to sequence labeling predictions.

A sentence-level representation $s_h$ is obtained as the weighted sum over all the value vectors in the sentence (Equation \ref{eq:sentence_rep}).
This is followed by two feed-forward layers: the first one is non-linear and projects the sentence representation onto a smaller feature space, while the second one is linear and outputs a scalar sentence-level score $o_h \in \mathbb{R}$ for each head $h$ (Equation \ref{eq:attn_2}).

To make a sentence prediction, the sentence scores need to be collected across all heads. The challenge arises as these $H$ scores (equal to the number of token labels) have to map to the number of sentence labels $S$, which are not always in direct correspondence. To solve this, we use the fact that many text classification tasks have a \textit{default label} that is common between the token and the sentence label sets -- for example, the \textit{neutral} label for sentiment analysis or the \textit{no-named-entity} label for NER. In our datasets, only two situations arise:

\begin{enumerate}
\item $\boldsymbol{H = S \text{: }}$ Each sentence label has a corresponding word-tag, and thus one head associated with it. Therefore, we can directly concatenate the sentence scores across all heads into a vector $\tilde{o} = [o_1; o_2; ...; o_H]$. An example of such a task is sentiment analysis, as the possible labels (positive, negative, and neutral) are the same for both sentences and tokens.
\item $\boldsymbol{H \neq S \text{ and } S = 2} \text{: }$ The sentence labels are binary, while the token labels are multi-class, therefore an appropriate correspondence between the heads and the two sentence labels needs to be found. We concatenate the score obtained for the default head $o_d$ (corresponding to the default label) with the maximum score obtained for the non-default heads $o_{nd}$: $\tilde{o} = [o_d; o_{nd}]$, where $d$ and $nd$ are the indices of the default and non-default heads, respectively, and $o_{nd} = \max\limits_{h \neq d}(o_h)$. Named entity recognition can be an example of such a task -- while there are many possible tags on the token level, we only detect the binary presence of any named entities on the sentence level.
\end{enumerate}

A probability distribution $\tilde y \in \mathbb{R}^{S}$ over the sentence labels is obtained by applying a softmax on the extracted scores: $\tilde y = softmax(\tilde{o})$. The most probable label is returned as the sentence-level prediction. 

In our model, the sentence-level scores are directly intertwined with the token-level predictions in order to improve performance on both levels. The attention weights are explicitly using the same predicted scores as the token-level output. Therefore, when the model learns to detect specific types of tokens, it will also assign more importance to those tokens in the corresponding attention heads. 
At the same time, when the model learns to attend to particular tokens for sentence classification, this will also help in identifying the correct labels for the token-level tagging task. By joining the two tasks, the architecture is able to share the information on both levels of granularity and achieve better results. In addition, this allows us to explicitly teach the model to focus on the same evidence as humans when performing text classification, leading to more explainable systems. 

In Figure \ref{fig:architecture}, we illustrate how this architecture -- to which we refer to as the multi-head attention labeler (MHAL) -- is applied on one input word to compute one attention head.

\subsection{Optimization objectives}
\label{sec:optimization}

Our model can be optimized both as a sentence classifier and as a sequence labeler using a cross-entropy loss. Both $L_{sent}$ and $L_{tok}$ minimize the summation over the negative log likelihood between the predicted sentence (or token) label distribution and the gold annotation:

\begin{align}
L_{sent} & = -\sum\limits_{s}\sum\limits_{j=1}^{S}{y^{(s)}_{j}\log(\tilde{y}^{(s)}_{j})} \label{eq:sent_class_obj} \\
L_{tok} & = -\sum\limits_{s}\sum\limits_{i=1}^{N}\sum\limits_{j=1}^{H}{t^{(s)}_{ij} \log(\tilde{t}^{(s)}_{ij})}
\end{align}

where $y^{(s)}_j$ and $t^{(s)}_{ij}$ are binary indicator variables specifying whether sentence $s$ truly is a sentence of label $j$ and token $t$ at position $i$ in sentence $s$ truly is a token of tag type $j$, respectively.

Recall that the sentence label distribution is based on the attention evidence scores, which represent, in turn, the token scores used for word-level classifications. 
If we train our model solely as a sentence classifier (by providing only sentence-level annotations), the network will also learn to label individual tokens. 
As all the parameters used by the token labeling component are also part of the sentence classifier, they will be optimized during the sentence-level training.
Moreover, the network will learn the important areas of a sentence, combining the scores from individual words to determine the overall sentence label. 
In this way, our model performs \textit{zero-shot sequence labeling}, a type of transductive transfer learning ~\cite{RuderMTL}. In addition, when both levels receive supervision, the token signal encourages the network to put more weight on the attention heads indicative of the correct labels.

We include an auxiliary \textit{attention loss} objective, based on \newcite{jointly}, which encourages the model to more closely connect the two labeling tasks on different granularity levels. In its original formulation, the loss could only operate over binary labels, whereas we extend it for general multi-class classification by imposing two conditions on the attention heads:

\begin{enumerate}
\item There should be at least one word of the same label as the ground-truth sentence. Intuitively, most of the focus should be on the words indicative of the sentence type.
\item There should be at least one word that has a default label. Even if the sentence has a non-default class, it should still contain at least one default word.
\end{enumerate}

While these conditions are not true for every text classification task, they are applicable in many settings and hold true for all the datasets that we experimented with. The two conditions can be formulated as a loss function and then optimized during training:

\begin{align}
\begin{split}\label{eq:attn_obj}
L_{attn} & = \sum\limits_{s}{\big(\max\limits_i(\tilde{t}^{(s)}_{i,h=l})} - 1\big)^2 + \sum\limits_{s}{\big(\max\limits_i(\tilde{t}^{(s)}_{i,h=d}) - 1\big)^2}
\end{split}
\end{align}

\noindent where $d$ is the default label, $l$ is the true sentence label, $\tilde{t}^{(s)}_{i}$ is the predicted token label distribution for word $i$ in sentence $s$, and thus $\tilde{t}^{(s)}_{i,h}$ is the predicted probability of word $i$ having label $h$.

Next, we propose a custom regularization term for the multi-head attention mechanism to motivate the network to learn a truly distinct representation sub-space for each of the query vectors $q_h$. As opposed to the keys and values, which are associated with different words, the queries $q_h$ encapsulate the essence of a certain tag. Therefore, these vectors need to capture the distinctive features of a particular label and how it is different from other labels. To push the network towards this goal, we introduce the term $R_q$ and calculate it as the average cosine similarity between every pair of queries $q_h$ and $q_i$, with $h \neq i$ (equation \ref{eq:reg_queries}). $R_q$ penalizes high similarity between any two query vectors and motivates the model to push them apart. Thus, this technique imposes a wider angle between the queries, encouraging the model to learn unique, diverse, and meaningful vector representations for the tags.

\begin{equation} \label{eq:reg_queries}
R_q = \frac{2}{H(H-1)}\sum\limits_{h = 1}^{H-1}{\sum\limits_{i > h}^{H}{\frac{q_h \cdot q_i}{\left\Vert q_h \right\Vert_2 \cdot \left\Vert q_i \right\Vert_2 } }}
\end{equation}

Lastly, we include an auxiliary objective for language modeling (LM) operating over characters and words, following the settings proposed by \newcite{rei-2017-semi}. The hidden representations from the forward and backward LSTMs are mapped to a new, non-linear space and used to predict the next word in the sequence, from a fixed smaller vocabulary. Recently, many NLP systems using multi-task learning include LM objectives along the core task to inject corpus-specific information into the model, as well as syntactic and semantic patterns ~\cite{semisup,Peters2,akbik-etal-2018-contextual,marvin-linzen-2018-targeted}. In our case, we include an LM loss to help the model learn general language features. 
While performing well on language modelling itself is not an objective, we expect it to provide improved biases and language-specific knowledge that would benefit performance. 

The final loss function $L$ is a weighted sum of all the objectives described above. Setting particular coefficients $\lambda$ allows us to investigate the effect of the different components as well as controlling the flow of the supervision signal and the importance of each auxiliary task: $L = \lambda_{sent} L_{sent} + \lambda_{tok} L_{tok}  + \lambda_{attn} L_{attn} + \lambda_{R_q} R_q + \lambda_{LM} L_{LM}$.

\section{Experiments}
\label{sec:experiments}

In this section, our goal is to test whether the proposed joint training architecture is able to \textbf{1)} transfer knowledge between words and sentences and improve on both text labeling tasks, \textbf{2)} learn to re-use the supervision signal received on the sentence-level to perform a word-level task, and \textbf{3)} use the auxiliary objectives and regularization loss to improve its performance. We perform three main experiments under different training regimes:

\begin{itemize}
\item \textbf{Fully supervised}: full annotations are provided both for sentences and words. The model has all the information needed to perform well at each separate level (i.e., in isolation). However, we are mainly interested in how performance changes as we train two related tasks together: does such a model take advantage of the joint learning regime and the supplemental labeled data?
\item \textbf{Semi-supervised}: some supervision signal is provided, but only for a subset of the words, while sentences are always receiving it in full. Under this setting, we determine the proportion of token annotation that is sufficient for the network to reach as good a performance as the fully supervised one. We check whether the (more instructed) sentence representations can pass unified, reusable knowledge about the entire sentence to its composing words.
\item \textbf{Unsupervised}: no word-level annotations are provided, but we test whether the model learns to perform sequence labeling, solely based on the sentence level signal, which is always provided in full (this is called \textit{zero-shot sequence labeling}). In other words, we train a sentence classifier and evaluate it as a sequence labeler. Under this setting, we aim to assess how much implicit knowledge a model can acquire about a low-level task (on words) solely by being trained on a higher-level task (on sentences). This zero-shot experiment is challenging: supervision signal is solely received at a higher, abstract sentence-level, while the task to be evaluated is at a lower, fine-grained token-level. If successful, this model will be able to perform sophisticated word-predictions solely based on the considerably cheaper sentence annotations.
\end{itemize}

\subsection{Data}
\label{sec:data}

To evaluate all the different properties of the model, we focus on three text classification datasets where annotations are available either for both individual words and full sentences or only for the words, but we can infer the ones for the sentences. We show some concrete examples in Table \ref{tab:task_examples}.

\textbf{SST}: The Stanford Sentiment Treebank ~\cite{socher-etal-2013-recursive} is a dataset of human-annotated movie reviews used for sentiment analysis. It contains not only sentence annotations for positive (P), negative (N), and neutral (O) reviews but also phrase-level annotations, which we converted to token labels by accounting for the minimum spans of tokens (up to length three) of a certain sentiment. Therefore, on SST we have three labels both at the sentence and at the word level.

\textbf{CoNLL03}: The CoNLL-2003 dataset ~\cite{CoNLL-2003} for named entity recognition (NER), contains five possible word-level tags, for person, organization, location, miscellaneous, or other, which is used for non-named entities. At the sentence-level, binary classification labels can be inferred based on whether the sentence contains any entities (annotated with label $\overline{\text{O}}$) or not (annotated with $\text{O}$).

\textbf{FCE}: The First Certificate in English ~\cite{yannakoudakis-etal-2011-new} is a dataset for fine-grained grammatical error detection. Ungrammatical words can contain five possible mistakes: in content, form, function, orthography, or other. There is also a sixth label for grammatical words. A sentence that contains at least one world-level mistake is ungrammatical overall. Therefore, a binary sentence classification task naturally occurs, as sentences can be grammatical (annotated with $\text{O}$) or not (annotated with $\overline{\text{O}}$).

All datasets are already tokenized and split into training, dev, and test sets. In Appendix~\ref{sec:appendix}, we provide some  statistics on the corpus (Table \ref{tab:corpora_stats}) and on the annotations available per split and label (Table \ref{tab:corpus_splits}).

\begin{table}[hb]
\centering
\resizebox{0.95\textwidth}{!}{
\begin{tabular}{lccccccccccc}
\toprule
\tabhead{Task} & \multicolumn{1}{c}{\tabhead{Sentence label}} & \multicolumn{10}{c}{\tabhead{\large{Sentences with word-level annotations}}}\\

\midrule

\multirow{2}{*}{\textbf{SST}}  & \multirow{2}{*}{P (positive)} & P & O & O & O & O & O & O & N & N & O \\
 &  & Good & acts & keep & it & from & being & a & total & rehash & . \\

\midrule

 \multirow{2}{*}{\textbf{CoNLL03}} & \multirow{2}{*}{$\overline{\text{O}}$ (has entity)} & O & O & O & LOC & O & PER & O & O & PER & O \\
 &  & New & talks & in & Chechnya & as & Lebed & waits & for & Yeltsin & . \\

\midrule

 \multirow{2}{*}{\textbf{FCE}} & \multirow{2}{*}{$\overline{\text{O}}$ (has error)} & O & CONT & FORM & O & O & O & FUNC & O & CONT & O \\
 &  & I & could & win & the & lottery & : & a & dream & too & ! \\
\bottomrule
\end{tabular}}
\caption{\label{tab:task_examples} Examples of text classification tasks with annotated labels for both sentences and words. We use $\text{O}$ and $\overline{\text{O}}$ to denote default and non-default sentence labels, respectively.}
\end{table}

\subsection{Hyperparameter settings}

We chose the best values for our  hyperparameters based on the performance on the development set (see Table \ref{tab:hyperparameter} in Appendix \ref{sec:appendix}). We perform each experiment with five different random seeds and report the average results. Following \newcite{attentionAll}, we also applied label smoothing \cite{label_smoothing} with $\epsilon = 0.15$ to increase the robustness to noise and regularize the label predictions during training. As evaluation metrics, we report (based on the task) the precision (P), accuracy (Acc), and micro-averaged F$_1$ score of all the labels and of all the non-default labels (denoted by a superscript $*$), as it is common in the multi-task learning literature ~\cite{changpinyo-etal-2018-multi,martinez-alonso-plank-2017-multitask}. For CoNLL03, we use the dedicated CoNLL evaluation script, which calculates F$_1$ on the entity level.

\subsection{Model variants}

We can optimize different variations of the architecture by changing the $\lambda$ weights in the loss and thereby choosing which components are active. We experiment with the following variations of the model:

\begin{itemize}
\item \textbf{MHAL-joint}: Corresponds to the fully supervised experiment, and is optimized both as a sentence classifier and a sequence labeler by setting $\lambda_{sent} = \lambda_{tok} = 1.0$, while all the other $\lambda$ values are $ 0.0$.
\item \textbf{MHAL-joint+}: 
In addition to training on both sentences and tokens ($\lambda_{sent} = \lambda_{tok} = 1.0$), the auxiliary objectives are also activated ($\lambda_{attn} = 0.01$,  $\lambda_{LM} = 0.1 $, and $R_q = 0.5$).
\item \textbf{MHAL-sent}: 
The model receives only sentence-level supervision  ($\lambda_{sent} = 1.0$) while all the other $\lambda$ values are set to $0.0$.
No supervision is provided on the token level, which means this model performs zero-shot sequence labeling.
\item \textbf{MHAL-sent+}: 
Receives supervision on the sentence level ($\lambda_{sent} = 1.0$), with the auxiliary objectives also activated ($\lambda_{attn} = 0.01$,  $\lambda_{LM} = 0.1 $, and $R_q = 0.5$). No supervision is provided on the token level ($\lambda_{tok} = 0.0$), so this model also performs zero-shot sequence labeling.
\end{itemize}

For comparison, we also evaluate two baseline models that do not connect the different hierarchical levels and specialize only on sentence classification or sequence labeling.
\begin{itemize}
\item \textbf{BiLSTM-sent}: Following the description and implementation of \newcite{yang-etal-2016-hierarchical}, we built one of the strongest neural sentence classifiers based on BiLSTMs and soft attention; we tuned the hyper-parameters based on the development set to achieve the best performance on our tasks.
\item \textbf{BiLSTM-tok}: Widely-used bidirectional LSTM architecture for sequence labeling, which has been applied to many tasks including part-of-speech tagging ~\cite{plank-etal-2016-multilingual} and named entity recognition ~\cite{panchendrarajan-amaresan-2018-bidirectional}. We also tuned the hyperparameters based on the development set in order to achieve the best results on each of the evaluation datasets.
\end{itemize}

\subsection{Results}

\textbf{Fully-supervised:} In this setting, we investigate whether training a joint model to solve the task on multiple levels provides a performance improvement over focusing only on one level. Table \ref{tab:multi_vs_single} compares the MHAL joint text classification performances to a BiLSTM attention-based sentence classifier and a BiLSTM sequence labeler. 
The results show that the multi-task models systematically outperform the single-task models across all tasks and datasets, emphasizing the effectiveness of sharing information between hierarchical levels. 
While additional annotation is required to train the multi-task models, the same input sentences are used in all cases, indicating that the benefits are coming directly from the model solving the task on multiple levels, as opposed to just from seeing more data examples.
Despite the sentence-level labels for CoNLL03 and FCE having been derived automatically from the existing token-level annotation, they still provide a performance improvement for sequence labeling, further showing the benefit of the multi-level architecture.
By teaching the model where to focus at the token level, the architecture is able to make better decisions on the sentence-level classification task. In addition, the sentence-level objective acts as a contextual regularizer and encourages the model to learn better compositional representations, thereby improving performance also on the token-level labeling task.

Comparing MHAL-joint against MHAL-joint+ with auxiliary objectives shows further improvements. The attention loss optimizes the model to make matching predictions between both levels, while the language modeling objective encourages the network to learn more informative word representations and composition functions. We also separately evaluated the regularization term, $R_q$, and found that it helps more on the sequence labeling tasks. This implies that using the intermediate per-head queries, the model indeed learns unique sub-space representations that help it assess the uncertainty of each word-tag pair and strengthen its labeling decision.

\begin{table*}[t]
\centering
\resizebox{0.9\textwidth}{!}{
\begin{tabular}{c|cc|cc|cc|cc|cc|cc}
\toprule
\multirow{3}{*}{\tabhead{}} & \multicolumn{6}{c|}{\tabhead{Sentence classification}} & \multicolumn{6}{c}{\tabhead{Sequence labeling}} \\ \cmidrule{2-7} \cmidrule{8-13}

& \multicolumn{2}{c|}{\tabhead{SST}} & \multicolumn{2}{c|}{\tabhead{CoNLL03}} & \multicolumn{2}{c|}{\tabhead{FCE}} & \multicolumn{2}{c|}{\tabhead{SST}} & \multicolumn{2}{c|}{\tabhead{CoNLL03}} & \multicolumn{2}{c}{\tabhead{FCE}} \\
% Sent. classification
& F$_{1}^{*}$ & Acc
& F$_{1}^{*}$ & F$_{1}$
& F$_{1}^{*}$ & F$_{1}$
% Seq. labeling
& P$^{*}$ & F$_{1}^{*}$
& P & F$_1$ 
& P$^{*}$ & F$_{0.5}^{*}$ \\
\midrule
BiLSTM-sent & 73.18 & 65.47 & 98.22 & 97.13 & 84.67 & 78.56 & - & - & - & - & - & - \\
BiLSTM-tok & - & - & - & - & - & - & 85.26 & 78.16 & 89.57 & 90.45 & 43.22 & 23.48\\
\midrule
MHAL-joint & 77.30 & 70.14 & \textbf{98.50} & \textbf{97.53} & 85.13 & 78.92 & 87.52 & 79.21 & 91.05 & 91.37 & 43.41 & 24.47 \\ 
$+$ reg. $R_q$ & 77.34 & 69.95 & 98.48 & 97.24 & 85.15 & 78.53 & \textbf{87.63} & 79.21 & \textbf{91.17} & 91.37 & \textbf{46.51} & 25.34 \\
MHAL-joint+ & \textbf{77.53} & \textbf{70.24} & 98.47 & 97.32 & \textbf{85.17} & \textbf{79.50} & 87.47 & \textbf{79.65} & 91.02 & \textbf{91.38} & 45.66 & \textbf{28.25}\\
\bottomrule
\end{tabular}}
\caption{Results on sentence classification and sequence labeling, comparing MHAL (which solves the tasks simultaneously by joining the two levels) with BiLSTM-sent and BiLSTM-tok, its equivalent single-task models. Note that metrics over the non-default labels are denoted by a superscript $*$.} \label{tab:multi_vs_single}
\end{table*}

\begin{figure*}[t]
\centering
\includegraphics[width=0.9\textwidth]{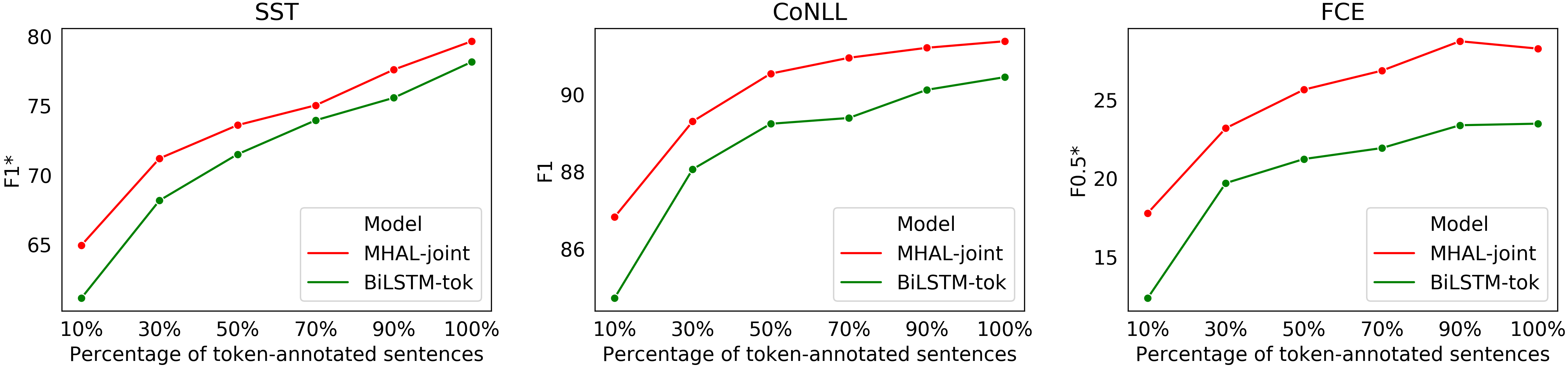}
\caption[]{Semi-supervised experiments for SST, CoNLL-2003, and FCE, comparing the sequence labeling performance of our multi-task model MHAL-joint+ with the single-task model BiLSTM-tok.}\label{fig:semisup}
\end{figure*}

\textbf{Semi-supervised:} We further experiment with MHAL-joint+, using the supervision signal of all sentences but varying the percentage \textit{p} of the word-level annotations. In Figure \ref{fig:semisup}, we present the sequence labeling results of the multi-task model, in comparison to the single-task BiLSTM-tok, gradually increasing \textit{p} to allow more tokens to guide learning. We observe that using only 30\%-50\% of the token-annotated data, our model already approaches the fully-supervised performance of the regular model (BiLSTM-tok-100\%), suggesting that the two tasks are positively influencing each other. General, abstract knowledge about the entire sentence meaning fluidly flows down to the words, while fine-grained word-level information is propagated up to the sentence, showing a beneficial transfer in both directions.

\begin{wraptable}{R}{0.45\textwidth}

\centering
\resizebox{0.45\textwidth}{!}{
\begin{tabular}{c|cc|cc|cc|cc|cc|cc}
\toprule
& \multicolumn{2}{c|}{\tabhead{SST}} & \multicolumn{2}{c|}{\tabhead{CoNLL03}} & \multicolumn{2}{c|}{\tabhead{FCE}} \\
& P$^{*}$ & F$_{1}^{*}$
& P & F$_1$ 
& P$^{*}$ & F$_{0.5}^{*}$ \\
\midrule
Random & 11.01 & 16.59 & 13.09 & 15.87 & 2.60 & 4.50 \\
\midrule
MHAL-sent & 8.67 & 10.62 & 13.18 & 16.47 & 2.46 & 4.24 \\
$+$ reg. $R_q$ & 14.70 & 22.07 & 13.99 & 17.48 & 2.33 & 4.03 \\
MHAL-sent+ & \textbf{21.60} & \textbf{26.93} & \textbf{24.02} & \textbf{25.51} & \textbf{4.03} & \textbf{5.64}\\
\bottomrule
\end{tabular}}
\caption{Zero-shot sequence labeling results.} \label{tab:zero_shot}

\end{wraptable}

\textbf{Unsupervised: } In Table \ref{tab:zero_shot}, we evaluate the architecture as a zero-shot sequence labeler, trained without any token-level supervision signal. 
In this setting, the model learns to label individual tokens by seeing only examples of sentence-level annotations. 
Because the multi-level attention is directly wired together with the sequence labeling output, the model is still able to learn in this difficult setting. 
This experiment also illustrates that the information does indeed flow from the sentence level down to individual tokens in this architecture.

As no other model can operate in this setting, we can only compare against a random baseline, in which the labels are assigned uniformly at random from the available set of labels. For datasets where the label distribution is very skewed, this can still be a difficult baseline to beat. While MHAL-sent only outperforms this baseline on the CoNLL03 dataset, MHAL-sent+ outperforms both on all datasets and metrics. These results show that our auxiliary losses introduce a necessary inductive bias and allow for better transfer of information from the sentence level to the tokens. We visualized the decisions computed inside the attention heads for different example sentences and provide them in Appendix \ref{sec:appendix} (Figures \ref{fig:zero_shot_visualisations_appendix} and \ref{fig:joint_visualisations}, respectively). We also observed that the choice of the metric based on which the stopping criterion is selected plays an important and interesting role in our zero-shot experiments (see Appendix \ref{sec:appendix_b} for details).

\section{Related work}
\label{sec:related_work}

Most methods for text classification (hierarchical or not) treat sentence classification and sequence labeling as completely separate tasks ~\cite{lample-etal-2016-neural,Huang,lei-etal-2018-multi,cui-zhang-2019-hierarchically}. More recently, training a model end-to-end for more than one language task has become increasingly popular ~\cite{YangTransferLF,Bert}, as well as using auxiliary objectives to inject useful inductive biases ~\cite{martinez-alonso-plank-2017-multitask,sogaard-goldberg-2016-deep,plank-etal-2016-multilingual,bingel-sogaard-2017-identifying}. Our work is similar in terms of motivation for the auxiliary objectives and multi-task training procedure. However, instead of learning to perform multiple tasks on the same level, we focus on performing the same task on multiple levels. By carefully designing the network and including specific auxiliary objectives, these levels are able to provide mutually beneficial information to each other. Other hierarchical multi-task systems, such as the models proposed by \newcite{hashimoto-etal-2017-joint} and \newcite{sanh}, solve each task at a different DNN layer, but their formulation does not follow a compositional linguistic motivation.

Our work is most similar to \newcite{jointly} and \newcite{rei-sogaard-2018-zero}, who described an architecture for supervising attention in a binary text classification setting. \newcite{barrett-etal-2018-sequence} also used a related model to guide the network to focus on similar areas as humans, based on human gaze recordings. We build on these ideas and describe a more general framework, extending it to both multiclass text classification and multiclass sequence labeling. An important part of our new architecture is based on attention mechanisms ~\cite{Bahdanau,luong-etal-2015-effective}, and, in particular, on the properties of multi-head attention ~\cite{attentionAll,li-etal-2019-information}. Other regularization techniques that explicitly encourage the learning of more diverse attention functions have been proposed by \newcite{li-etal-2018-multi}, who introduced a disagreement regularization term, and by \newcite{sparse_attention} and \newcite{correia-etal-2019-adaptively}, who proposed sparse attention for increased interpretability.

\section{Conclusion}
\label{sec:conclusion}

We investigated a novel neural architecture for natural language representations, which explicitly ties together predictions on multiple levels of granularity. The dynamically calculated weights in a multi-head attention component for composing sentence representations are also connected to token-level predictions, with each attention head focusing on detecting one particular label.
This model can then be optimized as either a sentence classifier or a token labeler, or jointly for both tasks, with information being shared between the two levels.
Supervision on the token labeling task also teaches the model where to assign more attention when composing sentence representations.
In return, the sentence-level objective acts as a contextual regularizer for the token labeler and encourages the model to predict token-level tags that cohere with the rest of the sentence.
We also introduce several auxiliary objectives that are further incentivizing the architecture to share information between the different levels and help this model get the most benefit from the available training data, increasing its efficiency.

We evaluated the proposed architecture on three different text classification tasks: sentiment analysis, named entity recognition, and grammatical error detection. 
The experiments showed that supervision on both levels of granularity consistently outperformed models that were optimized only on one level. 
This held true even for cases where the sentence-level labels could be automatically derived from token-level annotation, therefore requiring manually annotated labels only on one level.
The auxiliary objectives, designed to connect the predictions between the two tasks more closely, further improved model performance.
The semi-supervised experiments showed that this architecture can also be used with partial labeling -- with a 50-70\% reduction in token-annotated data, the model was able to get comparable results to the baseline architecture using the full dataset. 
Finally, we presented the first experiments for multi-class zero-shot sequence labeling, where the model needs to label tokens while only learning from the sentence-level annotation. 
As the architecture connects each attention head to a particular label, it was able to learn even in this challenging setting, with the auxiliary objectives being particularly beneficial.
The overall multi-level learning approach also has potential future applications in the area of neural network interpretability, as the model can be trained to focus on the same evidence as human users when classifying text and the resulting token-level decisions can be both measured and visualized.

\bibliographystyle{coling}
\bibliography{coling2020,anthology}

\begin{thebibliography}{}

\bibitem[\protect\citename{Akbik \bgroup et al.\egroup
  }2018]{akbik-etal-2018-contextual}
Alan Akbik, Duncan Blythe, and Roland Vollgraf.
\newblock 2018.
\newblock Contextual string embeddings for sequence labeling.
\newblock In {\em Proceedings of the 27th International Conference on
  Computational Linguistics}, pages 1638--1649, Santa Fe, New Mexico, USA,
  August. Association for Computational Linguistics.

\bibitem[\protect\citename{Bahdanau \bgroup et al.\egroup }2014]{Bahdanau}
Dzmitry Bahdanau, Kyunghyun Cho, and Yoshua Bengio.
\newblock 2014.
\newblock Neural machine translation by jointly learning to align and
  translate.
\newblock {\em arXiv}.

\bibitem[\protect\citename{Barrett \bgroup et al.\egroup
  }2018]{barrett-etal-2018-sequence}
Maria Barrett, Joachim Bingel, Nora Hollenstein, Marek Rei, and Anders
  S{\o}gaard.
\newblock 2018.
\newblock Sequence classification with human attention.
\newblock In {\em Proceedings of the 22nd Conference on Computational Natural
  Language Learning}, pages 302--312, Brussels, Belgium, October. Association
  for Computational Linguistics.

\bibitem[\protect\citename{Bingel and
  S{\o}gaard}2017]{bingel-sogaard-2017-identifying}
Joachim Bingel and Anders S{\o}gaard.
\newblock 2017.
\newblock Identifying beneficial task relations for multi-task learning in deep
  neural networks.
\newblock In {\em Proceedings of the 15th Conference of the {E}uropean Chapter
  of the Association for Computational Linguistics: Volume 2, Short Papers},
  pages 164--169, Valencia, Spain, April. Association for Computational
  Linguistics.

\bibitem[\protect\citename{Changpinyo \bgroup et al.\egroup
  }2018]{changpinyo-etal-2018-multi}
Soravit Changpinyo, Hexiang Hu, and Fei Sha.
\newblock 2018.
\newblock Multi-task learning for sequence tagging: An empirical study.
\newblock In {\em Proceedings of the 27th International Conference on
  Computational Linguistics}, pages 2965--2977, Santa Fe, New Mexico, USA,
  August. Association for Computational Linguistics.

\bibitem[\protect\citename{Correia \bgroup et al.\egroup
  }2019]{correia-etal-2019-adaptively}
Gon{\c{c}}alo~M. Correia, Vlad Niculae, and Andr{\'e} F.~T. Martins.
\newblock 2019.
\newblock Adaptively sparse transformers.
\newblock In {\em Proceedings of the 2019 Conference on Empirical Methods in
  Natural Language Processing and the 9th International Joint Conference on
  Natural Language Processing (EMNLP-IJCNLP)}, pages 2174--2184, Hong Kong,
  China, November. Association for Computational Linguistics.

\bibitem[\protect\citename{Cui and Zhang}2019]{cui-zhang-2019-hierarchically}
Leyang Cui and Yue Zhang.
\newblock 2019.
\newblock Hierarchically-refined label attention network for sequence labeling.
\newblock In {\em Proceedings of the 2019 Conference on Empirical Methods in
  Natural Language Processing and the 9th International Joint Conference on
  Natural Language Processing (EMNLP-IJCNLP)}, Hong Kong, China. Association
  for Computational Linguistics.

\bibitem[\protect\citename{Dai and Le}2015]{semisup}
Andrew~M. Dai and Quoc~V. Le.
\newblock 2015.
\newblock Semi-supervised sequence learning.
\newblock In {\em Proceedings of the 28th International Conference on Neural
  Information Processing Systems}, pages 3079--3087. MIT Press.

\bibitem[\protect\citename{Devlin \bgroup et al.\egroup }2018]{Bert}
Jacob Devlin, Ming-Wei Chang, Kenton Lee, and Kristina Toutanova.
\newblock 2018.
\newblock Bert: Pre-training of deep bidirectional transformers for language
  understanding.
\newblock {\em arXiv preprint arXiv:1810.04805}.

\bibitem[\protect\citename{Graves and Schmidhuber}2005]{GravesAndSchmidhuber}
Alex Graves and J\"{u}rgen Schmidhuber.
\newblock 2005.
\newblock Framewise phoneme classification with bidirectional lstm and other
  neural network architectures.
\newblock {\em NEURAL NETWORKS}, pages 5--6.

\bibitem[\protect\citename{Hashimoto \bgroup et al.\egroup
  }2017]{hashimoto-etal-2017-joint}
Kazuma Hashimoto, Caiming Xiong, Yoshimasa Tsuruoka, and Richard Socher.
\newblock 2017.
\newblock A joint many-task model: Growing a neural network for multiple {NLP}
  tasks.
\newblock In {\em Proceedings of the 2017 Conference on Empirical Methods in
  Natural Language Processing}, pages 1923--1933, Copenhagen, Denmark,
  September. Association for Computational Linguistics.

\bibitem[\protect\citename{Hornik}1991]{hornik}
Kurt Hornik.
\newblock 1991.
\newblock Approximation capabilities of multilayer feedforward networks.
\newblock {\em Neural Networks}, 4(2):251--257.

\bibitem[\protect\citename{Huang \bgroup et al.\egroup }2015]{Huang}
Zhiheng Huang, Wei Xu, and Kai Yu.
\newblock 2015.
\newblock Bidirectional lstm-crf models for sequence tagging.
\newblock {\em arXiv preprint arXiv:1508.01991}.

\bibitem[\protect\citename{Lample \bgroup et al.\egroup
  }2016]{lample-etal-2016-neural}
Guillaume Lample, Miguel Ballesteros, Sandeep Subramanian, Kazuya Kawakami, and
  Chris Dyer.
\newblock 2016.
\newblock Neural architectures for named entity recognition.
\newblock In {\em Proceedings of the 2016 Conference of the North {A}merican
  Chapter of the Association for Computational Linguistics: Human Language
  Technologies}, pages 260--270, San Diego, California, June. Association for
  Computational Linguistics.

\bibitem[\protect\citename{Lei \bgroup et al.\egroup
  }2018]{lei-etal-2018-multi}
Zeyang Lei, Yujiu Yang, Min Yang, and Yi~Liu.
\newblock 2018.
\newblock A multi-sentiment-resource enhanced attention network for sentiment
  classification.
\newblock In {\em Proceedings of the 56th Annual Meeting of the Association for
  Computational Linguistics (Volume 2: Short Papers)}, pages 758--763,
  Melbourne, Australia, July. Association for Computational Linguistics.

\bibitem[\protect\citename{Leshno \bgroup et al.\egroup }1993]{Moshe_ufa}
Moshe Leshno, Vladimir~Ya. Lin, Allan Pinkus, and Shimon Schocken.
\newblock 1993.
\newblock Original contribution: Multilayer feedforward networks with a
  nonpolynomial activation function can approximate any function.
\newblock {\em Neural Netw.}, 6:861–867, June.

\bibitem[\protect\citename{Li \bgroup et al.\egroup }2018]{li-etal-2018-multi}
Qian Li, Ziwei Li, Jin-Mao Wei, Yanhui Gu, Adam Jatowt, and Zhenglu Yang.
\newblock 2018.
\newblock A multi-attention based neural network with external knowledge for
  story ending predicting task.
\newblock In {\em Proceedings of the 27th International Conference on
  Computational Linguistics}, pages 1754--1762, Santa Fe, New Mexico, USA,
  August. Association for Computational Linguistics.

\bibitem[\protect\citename{Li \bgroup et al.\egroup
  }2019]{li-etal-2019-information}
Jian Li, Baosong Yang, Zi-Yi Dou, Xing Wang, Michael~R. Lyu, and Zhaopeng Tu.
\newblock 2019.
\newblock Information aggregation for multi-head attention with
  routing-by-agreement.
\newblock In {\em Proceedings of the 2019 Conference of the North {A}merican
  Chapter of the Association for Computational Linguistics: Human Language
  Technologies, Volume 1 (Long and Short Papers)}, pages 3566--3575,
  Minneapolis, Minnesota, June. Association for Computational Linguistics.

\bibitem[\protect\citename{Luong \bgroup et al.\egroup
  }2015]{luong-etal-2015-effective}
Thang Luong, Hieu Pham, and Christopher~D. Manning.
\newblock 2015.
\newblock Effective approaches to attention-based neural machine translation.
\newblock In {\em Proceedings of the 2015 Conference on Empirical Methods in
  Natural Language Processing}, pages 1412--1421, Lisbon, Portugal, September.
  Association for Computational Linguistics.

\bibitem[\protect\citename{Mart{\'\i}nez~Alonso and
  Plank}2017]{martinez-alonso-plank-2017-multitask}
H{\'e}ctor Mart{\'\i}nez~Alonso and Barbara Plank.
\newblock 2017.
\newblock When is multitask learning effective? semantic sequence prediction
  under varying data conditions.
\newblock In {\em Proceedings of the 15th Conference of the {E}uropean Chapter
  of the Association for Computational Linguistics: Volume 1, Long Papers},
  pages 44--53, Valencia, Spain, April. Association for Computational
  Linguistics.

\bibitem[\protect\citename{Marvin and Linzen}2018]{marvin-linzen-2018-targeted}
Rebecca Marvin and Tal Linzen.
\newblock 2018.
\newblock Targeted syntactic evaluation of language models.
\newblock In {\em Proceedings of the 2018 Conference on Empirical Methods in
  Natural Language Processing}, pages 1192--1202, Brussels, Belgium,
  October-November. Association for Computational Linguistics.

\bibitem[\protect\citename{Mikolov \bgroup et al.\egroup }2013]{Mikolov}
Tomas Mikolov, Ilya Sutskever, Kai Chen, Greg Corrado, and Jeffrey Dean.
\newblock 2013.
\newblock Distributed representations of words and phrases and their
  compositionality.
\newblock In {\em Proceedings of the 26th International Conference on Neural
  Information Processing Systems}, pages 3111--3119. Curran Associates Inc.

\bibitem[\protect\citename{Niculae and Blondel}2017]{sparse_attention}
Vlad Niculae and Mathieu Blondel.
\newblock 2017.
\newblock A regularized framework for sparse and structured neural attention.
\newblock In I.~Guyon, U.~V. Luxburg, S.~Bengio, H.~Wallach, R.~Fergus,
  S.~Vishwanathan, and R.~Garnett, editors, {\em Advances in Neural Information
  Processing Systems 30}, pages 3338--3348. Curran Associates, Inc.

\bibitem[\protect\citename{Panchendrarajan and
  Amaresan}2018]{panchendrarajan-amaresan-2018-bidirectional}
Rrubaa Panchendrarajan and Aravindh Amaresan.
\newblock 2018.
\newblock Bidirectional {LSTM}-{CRF} for named entity recognition.
\newblock In {\em Proceedings of the 32nd Pacific Asia Conference on Language,
  Information and Computation}, Hong Kong, 1{--}3 December. Association for
  Computational Linguistics.

\bibitem[\protect\citename{Pennington \bgroup et al.\egroup }2014]{glove}
Jeffrey Pennington, Richard Socher, and Christopher Manning.
\newblock 2014.
\newblock {G}love: Global vectors for word representation.
\newblock In {\em Proceedings of the Conference on Empirical Methods in Natural
  Language Processing}, pages 1532--1543. Association for Computational
  Linguistics.

\bibitem[\protect\citename{Peters \bgroup et al.\egroup }2017]{Peters2}
Matthew~E. Peters, Waleed Ammar, Chandra Bhagavatula, and Russell Power.
\newblock 2017.
\newblock Semi-supervised sequence tagging with bidirectional language models.
\newblock {\em arXiv preprint arXiv:1705.00108}.

\bibitem[\protect\citename{Piantadosi and Aslin}2016]{Piantadosi}
Steven Piantadosi and Richard Aslin.
\newblock 2016.
\newblock Compositional reasoning in early childhood.
\newblock {\em PloS one}, 11(9), Sep.

\bibitem[\protect\citename{Plank \bgroup et al.\egroup
  }2016]{plank-etal-2016-multilingual}
Barbara Plank, Anders S{\o}gaard, and Yoav Goldberg.
\newblock 2016.
\newblock Multilingual part-of-speech tagging with bidirectional long
  short-term memory models and auxiliary loss.
\newblock In {\em Proceedings of the 54th Annual Meeting of the Association for
  Computational Linguistics (Volume 2: Short Papers)}, pages 412--418, Berlin,
  Germany, August. Association for Computational Linguistics.

\bibitem[\protect\citename{Rei and S{\o}gaard}2018]{rei-sogaard-2018-zero}
Marek Rei and Anders S{\o}gaard.
\newblock 2018.
\newblock Zero-shot sequence labeling: Transferring knowledge from sentences to
  tokens.
\newblock In {\em Proceedings of the 2018 Conference of the North {A}merican
  Chapter of the Association for Computational Linguistics: Human Language
  Technologies, Volume 1 (Long Papers)}, pages 293--302, New Orleans,
  Louisiana, June. Association for Computational Linguistics.

\bibitem[\protect\citename{Rei and S{\o}gaard}2019]{jointly}
Marek Rei and Anders S{\o}gaard.
\newblock 2019.
\newblock Jointly learning to label sentences and tokens.
\newblock In {\em Proceedings of the 33rd National Conference on Artifical
  Intelligence}, pages 6916--6923. AAAI Press.

\bibitem[\protect\citename{Rei}2017]{rei-2017-semi}
Marek Rei.
\newblock 2017.
\newblock Semi-supervised multitask learning for sequence labeling.
\newblock In {\em Proceedings of the 55th Annual Meeting of the Association for
  Computational Linguistics (Volume 1: Long Papers)}, pages 2121--2130,
  Vancouver, Canada, July. Association for Computational Linguistics.

\bibitem[\protect\citename{Ruder}2017]{RuderMTL}
Sebastian Ruder.
\newblock 2017.
\newblock An overview of multi-task learning in deep neural networks.
\newblock {\em arXiv preprint arXiv:1706.05098}.

\bibitem[\protect\citename{Sandler}2018]{Sandler}
Wendy Sandler.
\newblock 2018.
\newblock The body as evidence for the nature of language.
\newblock {\em Frontiers in psychology}, 9:1782--1782, Oct.

\bibitem[\protect\citename{Sanh \bgroup et al.\egroup }2018]{sanh}
Victor Sanh, Thomas Wolf, and Sebastian Ruder.
\newblock 2018.
\newblock A hierarchical multi-task approach for learning embeddings from
  semantic tasks.
\newblock {\em arXiv preprint arXiv:1811.06031}.

\bibitem[\protect\citename{Shen and Lee}2016]{shen_and_lee_16}
Sheng{-}syun Shen and Hung{-}yi Lee.
\newblock 2016.
\newblock Neural attention models for sequence classification: Analysis and
  application to key term extraction and dialogue act detection.
\newblock {\em arXiv preprint arXiv:1604.00077}.

\bibitem[\protect\citename{Socher \bgroup et al.\egroup
  }2013]{socher-etal-2013-recursive}
Richard Socher, Alex Perelygin, Jean Wu, Jason Chuang, Christopher~D. Manning,
  Andrew Ng, and Christopher Potts.
\newblock 2013.
\newblock Recursive deep models for semantic compositionality over a sentiment
  treebank.
\newblock In {\em Proceedings of the 2013 Conference on Empirical Methods in
  Natural Language Processing}, pages 1631--1642, Seattle, Washington, USA,
  October. Association for Computational Linguistics.

\bibitem[\protect\citename{S{\o}gaard and
  Goldberg}2016]{sogaard-goldberg-2016-deep}
Anders S{\o}gaard and Yoav Goldberg.
\newblock 2016.
\newblock Deep multi-task learning with low level tasks supervised at lower
  layers.
\newblock In {\em Proceedings of the 54th Annual Meeting of the Association for
  Computational Linguistics (Volume 2: Short Papers)}, pages 231--235, Berlin,
  Germany, August. Association for Computational Linguistics.

\bibitem[\protect\citename{Spelke and Kinzler}2007]{spelke}
Elizabeth Spelke and Katherine Kinzler.
\newblock 2007.
\newblock Core knowledge.
\newblock {\em Developmental science}, 10(1):89--96.

\bibitem[\protect\citename{Szegedy \bgroup et al.\egroup
  }2016]{label_smoothing}
Christian Szegedy, Vincent Vanhoucke, Sergey Ioffe, Jonathon Shlens, and
  Zbigniew Wojna.
\newblock 2016.
\newblock Rethinking the inception architecture for computer vision.
\newblock In {\em Proceedings of the IEEE Conference on Computer Vision and
  Pattern Recognition}, pages 2818--2826.

\bibitem[\protect\citename{Tjong Kim~Sang and De~Meulder}2003]{CoNLL-2003}
Erik~F. Tjong Kim~Sang and Fien De~Meulder.
\newblock 2003.
\newblock Introduction to the conll-2003 shared task: Language-independent
  named entity recognition.
\newblock In {\em Proceedings of the 7th Conference on Natural Language
  Learning at HLT-NAACL 2003}, pages 142--147. Association for Computational
  Linguistics.

\bibitem[\protect\citename{Vaswani \bgroup et al.\egroup }2017]{attentionAll}
Ashish Vaswani, Noam Shazeer, Niki Parmar, Jakob Uszkoreit, Llion Jones,
  Aidan~N Gomez, \L~ukasz Kaiser, and Illia Polosukhin.
\newblock 2017.
\newblock Attention is all you need.
\newblock In {\em Advances in Neural Information Processing Systems 30}, pages
  5998--6008. Curran Associates, Inc.

\bibitem[\protect\citename{Wagner \bgroup et al.\egroup }2011]{Wagner2011}
Richard~K. Wagner, Cynthia~S. Puranik, Barbara Foorman, Elizabeth Foster,
  Laura~Gehron Wilson, Erika Tschinkel, and Patricia~Thatcher Kantor.
\newblock 2011.
\newblock Modeling the development of written language.
\newblock {\em Reading and writing}, 24(2):203--220, Feb.

\bibitem[\protect\citename{Yang \bgroup et al.\egroup
  }2016]{yang-etal-2016-hierarchical}
Zichao Yang, Diyi Yang, Chris Dyer, Xiaodong He, Alex Smola, and Eduard Hovy.
\newblock 2016.
\newblock Hierarchical attention networks for document classification.
\newblock In {\em Proceedings of the 2016 Conference of the North {A}merican
  Chapter of the Association for Computational Linguistics: Human Language
  Technologies}, pages 1480--1489, San Diego, California, June. Association for
  Computational Linguistics.

\bibitem[\protect\citename{Yang \bgroup et al.\egroup }2017]{YangTransferLF}
Zhilin Yang, Ruslan~R. Salakhutdinov, and William~W. Cohen.
\newblock 2017.
\newblock Transfer learning for sequence tagging with hierarchical recurrent
  networks.
\newblock {\em arXiv preprint arXiv:1703.06345}.

\bibitem[\protect\citename{Yannakoudakis \bgroup et al.\egroup
  }2011]{yannakoudakis-etal-2011-new}
Helen Yannakoudakis, Ted Briscoe, and Ben Medlock.
\newblock 2011.
\newblock A new dataset and method for automatically grading {ESOL} texts.
\newblock In {\em Proceedings of the 49th Annual Meeting of the Association for
  Computational Linguistics: Human Language Technologies}, pages 180--189,
  Portland, Oregon, USA, June. Association for Computational Linguistics.

\end{thebibliography}

\clearpage
\appendix

\section{Appendix A. Dataset statistics, hyperparameters and extra visualizations}
\label{sec:appendix}

\begin{table*}[ht]
\centering
\resizebox{0.7\textwidth}{!}{
\begin{tabular}{l|cc|cc|cc|cc}
\toprule
\multirow{2}{*}{\tabhead{Dataset}} & \multicolumn{2}{c|}{\tabhead{No. labels}} & \multicolumn{2}{c|}{\tabhead{Prop. O}} & \multicolumn{2}{c|}{\tabhead{Full Entropy}} & \multicolumn{2}{c}{\tabhead{Non-O Entropy}}\\
 & sent & tok & sent & tok & sent & tok & sent & tok \\
\midrule
SST & 3 & 3 & 0.19 & 0.78 & 1.509 & 0.961 & 0.999 & 0.956 \\
CoNLL-2003 & 2 & 5 & 0.20 & 0.83 & 0.731 & 0.979 & 0.263 & 1.929 \\
FCE & 2 & 6 & 0.37 & 0.89 & 0.952 & 0.775 & 0.421 & 2.288 \\
\bottomrule
\end{tabular}}
\caption[]{We list, for sentences and tokens: number of unique labels, proportion of default labels (O), entropy of the label distribution, and entropy of the non-default label distribution (using $\log_{2}$).} \label{tab:corpora_stats}
\end{table*}

\begin{table*}[ht]
\centering
\resizebox{0.65\textwidth}{!}{
\begin{tabular}{c|c|ccc|ccc}
\toprule
\multirow{2}{*}{\tabhead{Dataset}} & \multirow{2}{*}{\tabhead{Label}} & \multicolumn{3}{c|}{\tabhead{Number of sentences}} & \multicolumn{3}{c}{\tabhead{Number of tokens}}\\
\cmidrule(l){3-5} \cmidrule(l){6-8}
& & Train & Dev & Test & Train & Dev & Test \\

\midrule
\midrule

\multirow{4}{*}{SST}

& O & 1,624 & 229 & 389 & 128,156 & 16,684 & 33,128 \\ 
& N & 3,310 & 428 & 912 & 13,384 & 1,740 & 3,488 \\ 
& P & 3,610 & 444 & 909 & 22,026 & 2,850 & 5,789 \\ 
\cmidrule(l){2-8}
& Total & 8,544 & 1,101 & 2,210 & 163,566 & 21,274 & 42,405 \\

\midrule
\midrule

\multirow{6}{*}{CoNLL-2003}
& O & 2,909 & 645 & 697 & 169,578 & 42,759 & 38,323 \\ 
& LOC & \multirow{4}{*}{11,132} & \multirow{4}{*}{2,605} & \multirow{4}{*}{2,756} & 8,297 & 2,094 & 1,925 \\ 
& MISC &  &  &  & 4,593 & 1,268 & 918 \\ 
& ORG &  &  &  & 10,025 & 2,092 & 2,496 \\ 
& PER &  &  &  & 11,128 & 3,149 & 2,773 \\ 
\cmidrule(l){2-8}
& Total & 14,041 & 3,250 & 3,453 & 203,621 & 51,362 & 46,435 \\

\midrule
\midrule

\multirow{7}{*}{FCE} 
& O & 10,718 & 824 & 900 & 396,479 & 30,188 & 35,525 \\ 
& CONTENT & \multirow{5}{*}{17,836} & \multirow{5}{*}{1,384} & \multirow{5}{*}{1,806} & 7,194 & 527 & 673 \\ 
& FORM &  &  &  & 8,174 & 621 & 850 \\ 
& FUNC &  &  &  & 11,084 & 888 & 1,194 \\ 
& ORTH &  &  &  & 12,655 & 1,126 & 1,429 \\ 
& OTHER &  &  &  & 11,415 & 861 & 1,116 \\ 
\cmidrule(l){2-8}
& Total & 28,554 & 2,208 & 2,706 & 447,001 & 34,211 & 40,787 \\

\bottomrule
\end{tabular}}
\caption[]{Statistics of the labeled sentences and tokens, separated by the train, dev or test split.}
\label{tab:corpus_splits}
\end{table*}

\begin{table*}[ht]
\centering
\resizebox{0.85\textwidth}{!}{
\begin{tabular}{lcp{8.5cm}}
\toprule
\tabhead{Hyperparameter} & \tabhead{Value} & \tabhead{Description} \\ 
\midrule
word embedding size & 300 & Size of the word embeddings. \\ 
char embedding size & 100 & Size of the character embeddings. \\ 
word recurrent size  & 300 & Size of the word-level BiLSTM hidden layers. \\ 
char recurrent size & 100 & Size of the character-level BiLSTM hidden layers. \\
word hidden layer size & 50 & Compact word vector size, applied after the last BiLSTM. \\  
char hidden layer size  & 50 & Char representation size, applied before concatenation. \\ 
attention evidence size  & 100 & Layer size for predicting attention weights. \\ 
hidden layer size & 200 & Final hidden layer size, right before word-level predictions. \\ 
max batch size & 32 & Number of sentences taken for training. \\ 
epochs & 200 & Maximum number of epochs to run the experiment for. \\ 
stop if no improvement & 7 & Stop if there has been no improvement for this many epochs. \\ 
learning rate & 1.0 & The learning rate used in AdaDelta. \\ 
decay & 0.9 & Learning rate decay used in AdaDelta. \\ 
input dropout & 0.5 & Value of the dropout applied after the LSTMs. \\ 
attention dropout & 0.5 & Value of the dropout applied on the attention mechanism. \\ 
LM max vocab size & 7500 & Max vocabulary size for the language modeling objective. \\ 
smoothing epsilon & 0.15 & The value of the epsilon in label smoothing. \\ 
stopping criterion & F${_{1\mu}^*}$ & The development metric used as the stopping criterion. \\ 
optimization algorithm & AdaDelta & Optimization algorithm used. \\ 
initializer & Glorot & Method for random initialization. \\ 
\bottomrule
\end{tabular}}
\caption[]{Hyperparameter settings for all of our MHAL models.}
\label{tab:hyperparameter}
\end{table*}

\begin{figure*}[ht]
\centering
\includegraphics[width=0.6\textwidth]{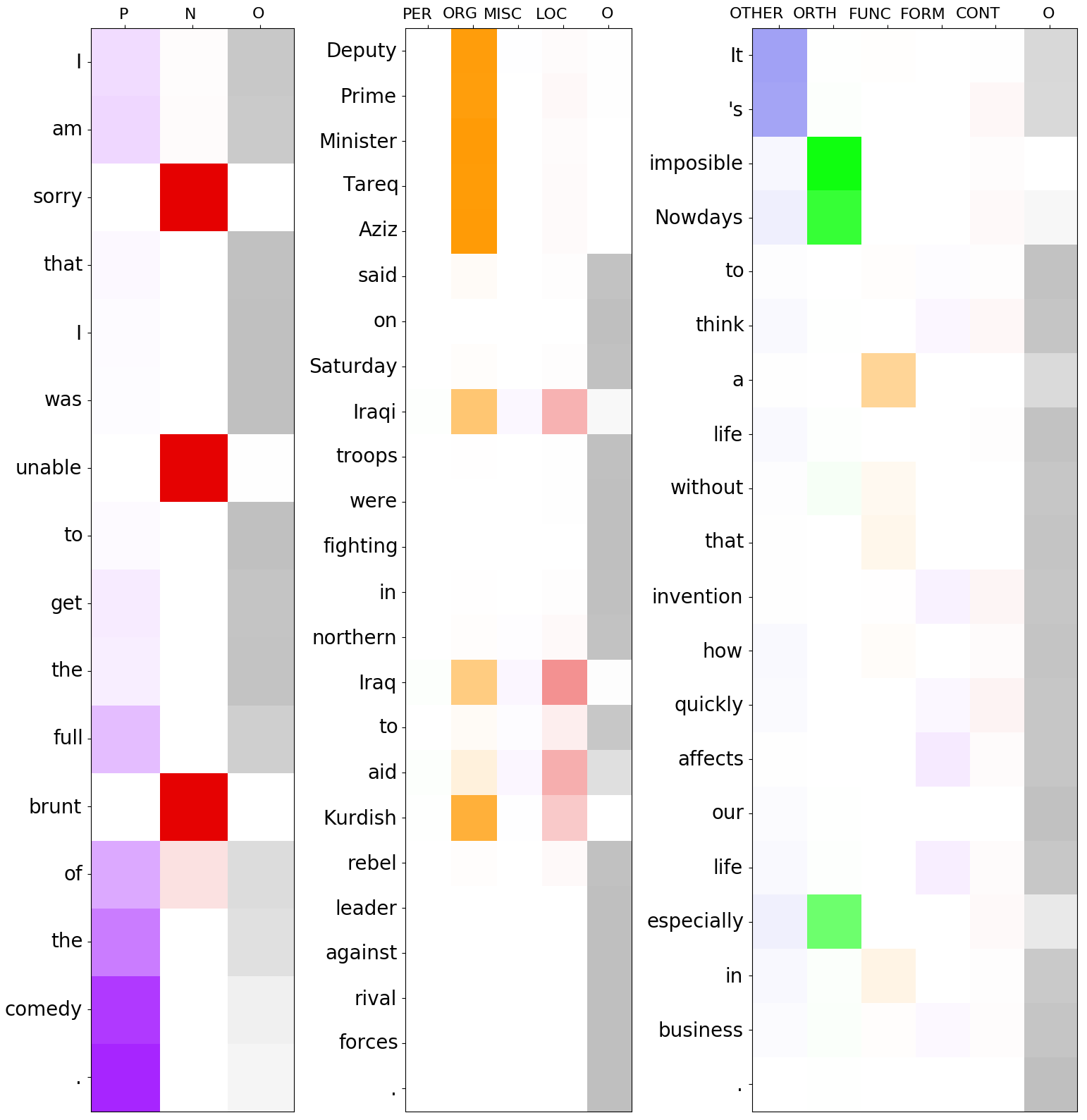}
\caption{Attention evidence scores, normalized acrossed heads, assigned by MHAL-sent for the words in three sentences from the SST (leftmost), CoNLL-2003 (middle), and FCE (rightmost) datasets.}\label{fig:zero_shot_visualisations_appendix}
\end{figure*}

\begin{figure*}[hb]
\centering
\includegraphics[width=0.6\textwidth]{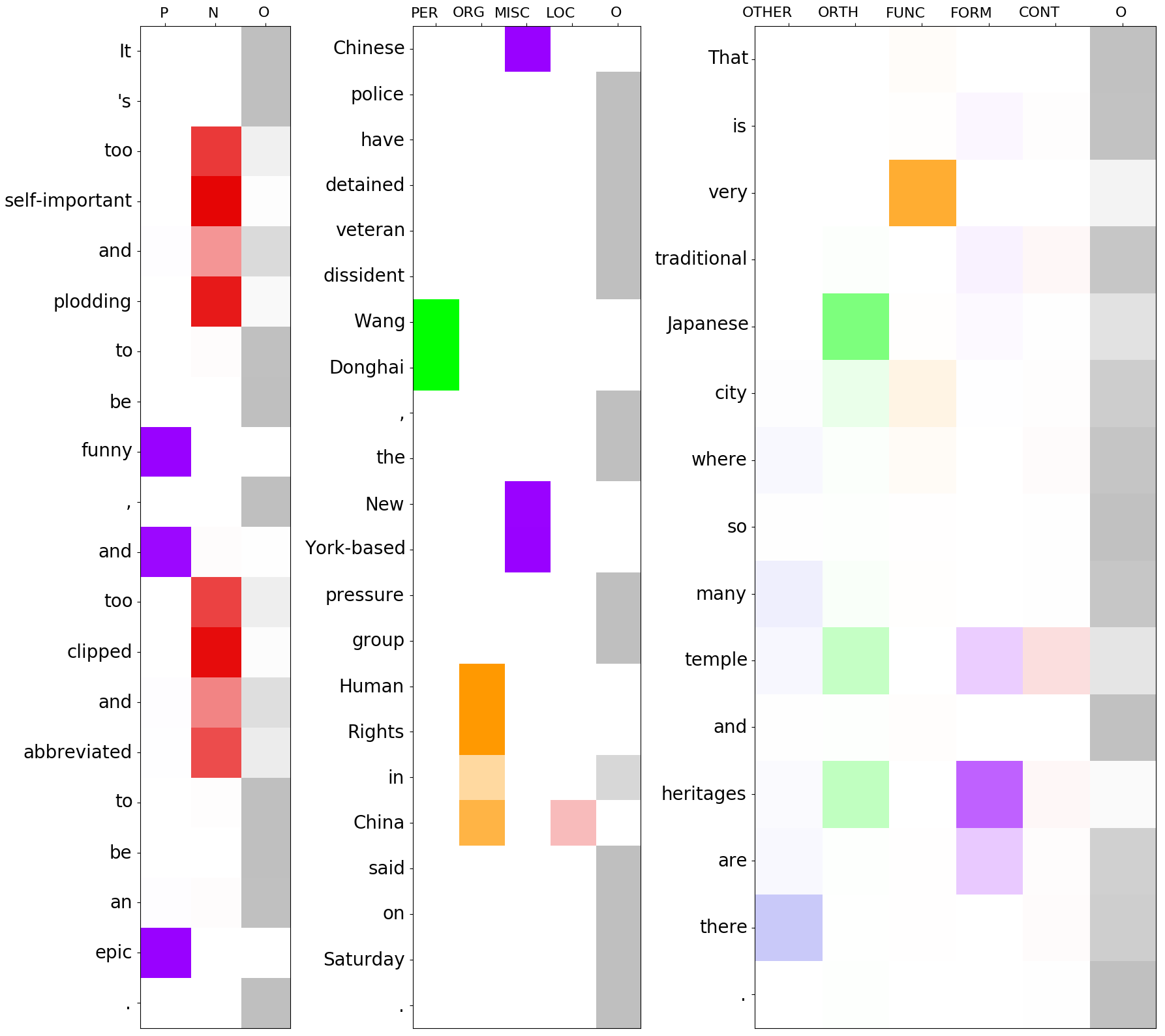}
\caption{Attention evidence scores, normalized acrossed heads, assigned by MHAL-joint+ for the words in three sentences from the SST (leftmost), CoNLL-2003 (middle), and FCE (rightmost) datasets.}\label{fig:joint_visualisations}
\end{figure*}

\section{Appendix B. Stopping criterion: results and discussion}
\label{sec:appendix_b}

\begin{table}[ht]
\centering
\resizebox{0.9\textwidth}{!}{
\begin{tabular}{l|cccc|cccc|cccc}
\toprule
\multirow{2}{*}{\tabhead{Dev metric}} & \multicolumn{4}{c|}{\tabhead{SST}} & \multicolumn{4}{c|}{\tabhead{CoNLL-2003}} & \multicolumn{4}{c}{\tabhead{FCE}} \\

& P$^{*}$ & R$^{*}$ & F$_{1}^{*}$ & S-Acc
& P & R & F$_1$ & S-F$_1$
& P$^{*}$ & R$^{*}$ & F$_{0.5}^{*}$ & S-F$_1$ \\
\midrule

S-F$_{1\mu}^{*}$ &  21.60 & 39.78 & 26.93 & \textbf{71.08} & 24.02 & 27.23 & 25.51 & 96.80 & \textbf{4.03} & 12.00 & 5.64 & 77.90 \\
F$_{1\mu}^{*}$ & 23.21 & 32.97 & 27.24 & 68.64 & 20.03 & 24.25 & 21.79 & 93.05 & 3.56 & \textbf{18.52} & 5.96 & 75.83 \\
(S-F$_{1\mu}^{*} + $F$_{1\mu}^{*}) / 2$ & \textbf{23.34} & \textbf{47.00} & \textbf{30.22} & 70.92 & \textbf{24.10} & \textbf{28.02} & \textbf{25.92} & \textbf{96.90} & 3.85 & 17.12 & \textbf{6.28} & \textbf{78.03} \\ 
\bottomrule
\end{tabular}}
\caption[]{The effect of the metric used in early stopping during the training of our zero-shot sequence labeler. Metrics prefixed by \textit{S-} represent sentence-level results.}
\label{tab:stopping_criterion_effect}
\end{table}

During model training, we measure performance on the development set and apply one of the two stopping criteria:
\textbf{1.} the sentence-level classification performance (S-F$_{1\mu}^{*}$), adopted by all models that do not receive any token-level annotation, such as MHAL-sent;
\textbf{2.} the token-level classification performance (F$_{1\mu}^{*}$), adopted by all models that receive some token annotation, such as MHAL-joint.

We observed that, even in the case of MHAL-sent, stopping based on the token performance improves the word-level predictions at test time, but usually hurts the sentence predictions. However, as suggested by the results in Table \ref{tab:stopping_criterion_effect}, stopping based on the average of these two metrics generally improves both the token and the sentence predictions. 

The network usually takes more time to reach the common optimal point when we include the token-based stopping criterion. Sentence classification converges faster than sequence labeling -- being predicted at a higher layer in the network hierarchy, it accumulates more information and thus builds solid abstractions, while having fewer unique instances to learn from. For these reasons, the network falls into a local minimum when guided by the sentence-level performance. However, choosing tokens as a stopping criterion requires annotated development data, which would not comply with the framing of our zero-shot learning experiment. Nevertheless, reporting this finding emphasizes that the stopping criterion requires careful consideration -- it is responsible for choosing the best performing model used during testing and for driving the application of the learning rate decay. Several performance percentage points could be gained by carefully selecting the stopping metric.

\end{document}